\documentclass[conference]{IEEEtran}
\IEEEoverridecommandlockouts
\usepackage{cite}
\usepackage{amsmath,amssymb,amsfonts}
\usepackage{algorithmic}
\usepackage{graphicx}
\usepackage{textcomp}
\usepackage{xcolor}
\usepackage{bbm}
\usepackage{makecell}
\usepackage{caption}
\usepackage{subcaption}
\def\BibTeX{{\rm B\kern-.05em{\sc i\kern-.025em b}\kern-.08em
    T\kern-.1667em\lower.7ex\hbox{E}\kern-.125emX}}
\begin{document}

\title{SE-PSNet: Silhouette-based Enhancement Feature for Panoptic Segmentation Network}

\author{Shuo-En Chang\textsuperscript{1}, Yi-Cheng Yang\textsuperscript{2}, En-Ting Lin\textsuperscript{1}, Pei-Yung Hsiao\textsuperscript{3}, \textit{Member, IEEE}, and Li-Chen Fu\textsuperscript{1}, \textit{Fellow, IEEE} \\
\IEEEauthorblockA{
\textsuperscript{1}Department of Computer Science and Information Engineering, National Taiwan University, Taipei, Taiwan \\
\textsuperscript{2}Graduate Institute of Networking and Multimedia, National Taiwan University, Taipei, Taiwan \\
\textsuperscript{3}Department of Electrical Engineering, National University of Kaohsiung, Kaohsiung, Taiwan\\
\{r08922a02, r09944023, b06902023, lichen\}@ntu.edu.tw, pyhsiao@nuk.edu.tw}
}

\maketitle

\begin{abstract}
Recently, there has been a panoptic segmentation task combining semantic and instance segmentation, in which the goal is to classify each pixel with the corresponding instance ID. In this work, we propose a solution to tackle the panoptic segmentation task. The overall structure combines the bottom-up method and the top-down method. Therefore, not only can there be better performance, but also the execution speed can be maintained. The network mainly pays attention to the quality of the mask. In the previous work, we can see that the uneven contour of the object is more likely to appear, resulting in low-quality prediction. Accordingly, we propose enhancement features and corresponding loss functions for the silhouette of objects and backgrounds to improve the mask. Meanwhile, we use the new proposed confidence score to solve the occlusion problem and make the network tend to use higher quality masks as prediction results. To verify our research, we used the COCO dataset and CityScapes dataset to do experiments and obtained competitive results with fast inference time. 
\end{abstract}


\section{Introduction}
Panoptic segmentation is a task combining semantic and instance segmentation. It can help the computer perceives daily life more correctly. For instance, in the field of autonomous driving, computers should recognize sidewalks as well as pedestrians. Several methods for panoptic segmentation have been proposed in the literature. Whether it is a proposed-based or proposed-free method, most of them will separate the task into semantic segmentation and instance segmentation. Then, do post-processing to combine both and generate the panoptic segmentation prediction. Observing the results of previous experiments, we found that the quality of the mask in instance segmentation will significantly affect the panoptic prediction. Although past research can achieve great performance on panoptic segmentation, the distinct silhouette segmentation was not what they consider. As a result, it will make the mask become lousy quality. Another issue is that when we sort the prediction results only according to the class confidence score, this will cause the performance to drop down when the occlusion happens. Small things usually have lower class confidence scores. However, according to the definition of panoptic segmentation, each pixel can only belong to one object. Accordingly, the small object cannot be capture.

To overcome the above issues, we propose a novel panoptic segmentation framework called Silhouette-based Enhancement Feature for Panoptic Segmentation Network (SE-PSNet). It adapted two branches, namely, semantic segmentation branch and instance segmentation branch. The semantic segmentation branch will produce the mask prediction in a fully convolution fashion with silhouette-based enhancement features. The proposed instance branch aims to generate bases and the attention map for each instance. Bases can be viewed as roughly mask predictions for the entire image, and combining with the individual attention map can generate the refined instance mask prediction. Silhouette-based enhancement features will help to improve the mask quality in both processes of the instance branch. Furthermore, we do not use class confidence score but using mask quality score to achieve a better result.

\section{Related Work}

\subsection{Panoptic Segmentation with Proposed-based}
The proposed-based methods are also known as top-down methods. That is because they will first detect the place of the object and do the segmentation later. Many works adapted Mask-RCNN\cite{he2017maskrcnn} as their instance segmentation branch parallel with a lightweight semantic segmentation branch using the shared backbone. Some of the methods will be introduced following. 
TASCNet\cite{li2018tascnet} proposed a consistency loss to do cross-task constraint, aiming to ensure alignment between thing prediction and stuff prediction.
AUNet\cite{li2019aunet} adds two attention sources to the stuff branch: from the RPN layer and foreground segmentation mask, which can provide object-level and pixel-level attention, respectively.
Panoptic FPN\cite{kirillov2019panopticfpn} endows Mask R-CNN\cite {he2017maskrcnn} with a lightweight semantic segmentation branch using a shared Feature Pyramid Network\cite {lin2017fpnnet} backbone, which profoundly affected the latter method. 
UPSNet\cite{Xiong2019UPSNetAU} introduced a parameter-free panoptic head that solves the panoptic segmentation via pixel-wise classification.
OANet\cite{liu2019oanet} uses a spatial ranking module to solve the multiple assignments for one pixel, also known as the occlusion problem.
OCFusion\cite{lazarow2020ocfusion} is another method that aims to solve the occlusion problem with an additional head predicting the occlusion relationship between two instances.
AdaptIS\cite{sofiiuk2019adaptis} uses an image and point proposal as input and outputs a mask of an object corresponding to that point. Different from other methods, it will generate class-agnostic instance segmentation and can be combined with a standard semantic segmentation pipeline.
SOGNet\cite{yang2020sognet} performs relational embedding, which can explicitly encode overlap relations without direct supervision on them.
EPSNet\cite{chang2020epsnet} proposed a cross-layer attention fusion module to capture the long-range dependencies between different scales feature maps.
Unifying\cite{Li2020UnifyingTA} uses a novel pairwise instance affinity operation with the panoptic matching loss, which enables end-to-end training and heuristics-free inference.
CondInst\cite{tian2021conditional} uses dynamic generates kernel's weight of the mask head to get the mask predictions.
EfficientPS\cite{mohan2021efficientps} proposed a new strong panoptic backbone and a panoptic fusion module to yield the final panoptic segmentation output.

Although the top-down methods often can get better performance, they usually need longer computation time since it needs to detect the rough object first. In addition, most of the situation, inconsistency will happen between stuff head and instance head. Adding the silhouette-based enhancement features can help to reduce this situation since it belongs to their shared features.

\subsection{Panoptic Segmentation with Proposed-free}
Unlike the previous group, the proposed-free method does not need to detect the rough object first. 
DeeperLab\cite{yang2019deeperlab} is the first bottom-up approach. They adopted an encoder-decoder topology, which follows the design of DeepLab\cite{chen2018deeplabv3+}, to predict instance keypoints multi-range offset heatmaps, then gather them into class-agnostic instance segmentation. 
Panoptic-DeepLab\cite{Cheng2020PanopticDeepLabAS} was also built on DeepLab\cite{chen2018deeplabv3+} and proposed the dual-ASPP and dual-decoder structures for the semantic branch and instance branch. 
SSAP\cite{gao2019ssap} proposed grouping pixels based on a pixel-pair affinity pyramid and incorporating a novel cascaded graph partition module to generate instances efficiently. 
Panoptic FCN\cite{li2021panopticfcn} proposed kernel generator and kernel fusion to generate the kernel weight for each object instance and semantic category. 

Even though proposed-free method have faster inference speed, most of there performance still exist a large gap between proposed-based method. In our work, we adopt a proposed-based method with an one-stage anchor-free instance segmentation framework, which can get a balance between the performance and computation time.

\subsection{Boundary Learning in Deep Learning}
The boundary is a critical feature that can be used in the real world. Since the edge and mask are complementary, a great boundary prediction can help us improve segmentation performance. In the literature, some of them use boundaries to guide the prediction of segmentation. Others directly predict the contour as the segmentation result. 
Deng \textit{et al.}\cite{deng2018boundaries} proposed a simple convolutional encoder-decoder network to predict crisp boundaries.
Edgenet\cite{han2021edgenet} uses a class-aware edge loss, which can improve the classification result of those pixels near the semantic segmentation boundaries.
Zimmermann \textit{et al.}\cite{zimmermann2019faster} uses classical edge detection filters applied on each instance mask, encouraging a better prediction near the instance boundaries.
BMask R-CNN\cite{cheng2020boundary} first predicts instance-level boundaries separately from the instance mask, then uses them to guide the mask learning with fusion fashion.
PolarMask represents a mask by its contour. Thus, it only needs to predict one center and rays emitted from the center on the polar coordinate to generate the instance mask.
DeepSnake\cite{peng2020deepsnake} performs instance segmentation by deforming an initial contour to match object boundary with proposed circular convolution.
Boundary IoU\cite{cheng2021boundary} was a new segmentation evaluation measure focused on boundary quality.

In our work, we use the boundary feature as the enhancement feature. This can help the network get the finer segmentation, especially when different instance occlusion. Furthermore, it can help the backbone network learn generalized features for the stuff branch and instance branch.

\section{Network Architecture Design}
Our method consists of six major components including (1) shared backbone, (2) stuff head, (3) box head, (4) basis head, (5) mask head, and (6) confidence head. 

\subsection{Backbone Network}
\begin{figure}
    \centering
    \includegraphics[width=\linewidth]{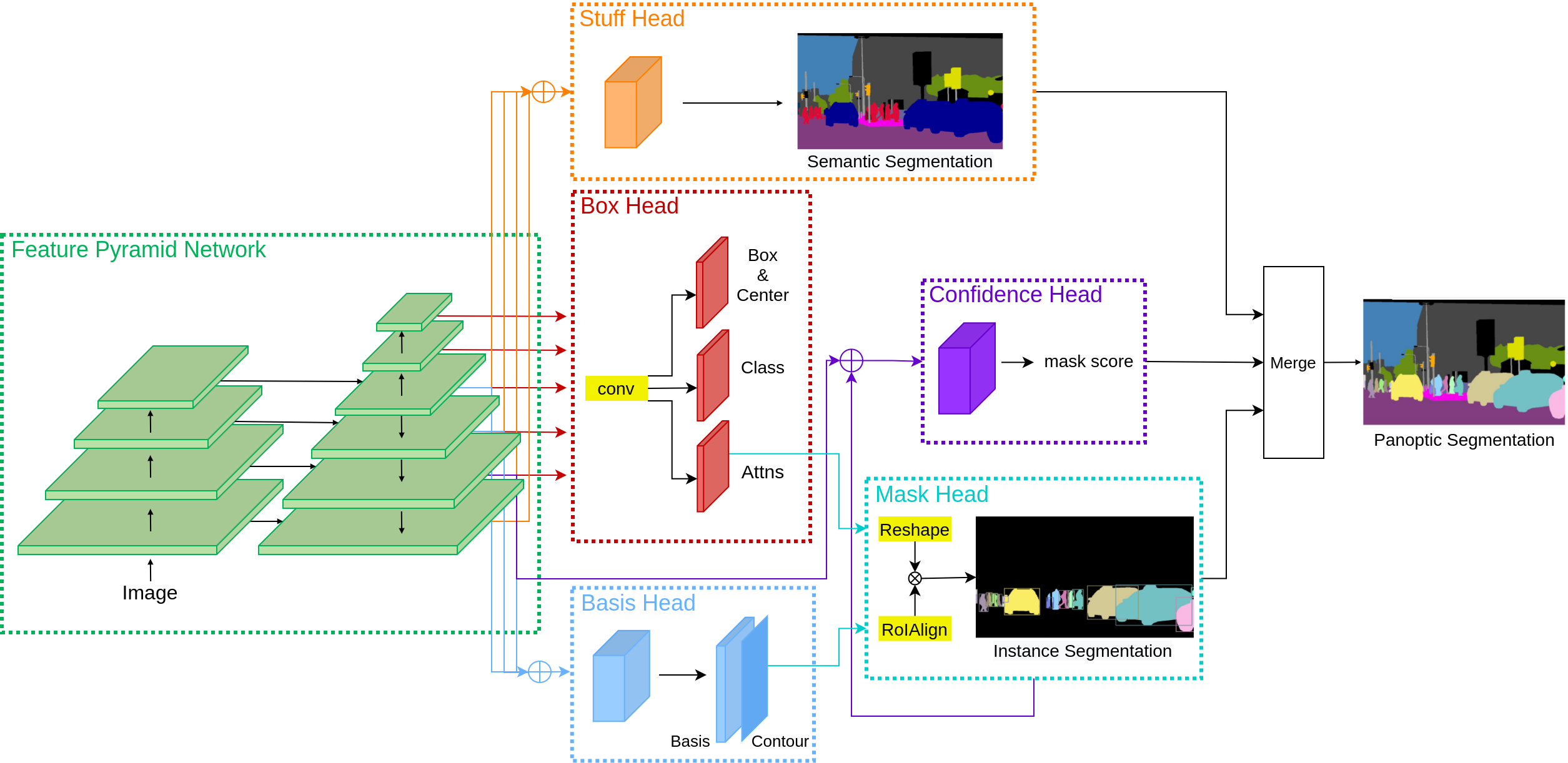}
    \caption{The architecture of SE-PSNet}
    \label{fig:overview}
\end{figure}
We used Resnet-101 as our backbone network and pretrained it on ImageNet\cite{deng2009imagenet}. 
As shown in Fig. \ref{fig:overview}, FPN is a top-down architecture with lateral connections, which will generate pyramid features $P_2$ to $P_7$ in different scales. We send $P_3$ to $P_7$ separately into the box head. On the other hand, $P_3$ to $P_5$ is used in the basis head, and $P_2$ to $P_5$ is used in the stuff head. Detail will be mentioned in the following section.

\subsection{Silhouette Features}

\begin{figure}[b]
    \centering
    \begin{subfigure}{0.24\textwidth}
        \includegraphics[width=\linewidth]{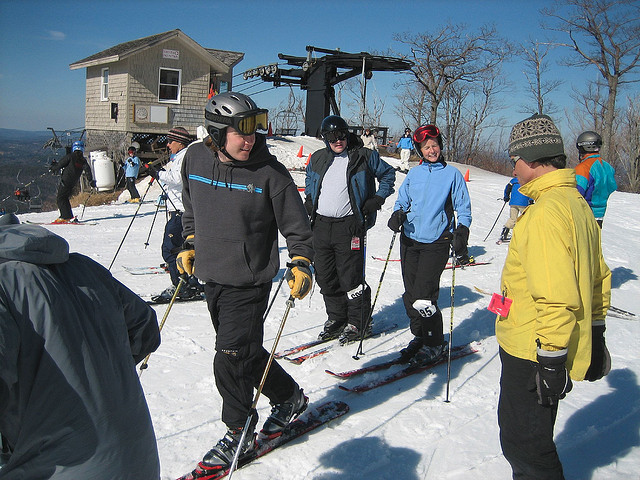}
        \caption{Example image}
        \label{fig:contour_img}
    \end{subfigure} \hfil
    \begin{subfigure}{0.24\textwidth}
        \includegraphics[width=\linewidth]{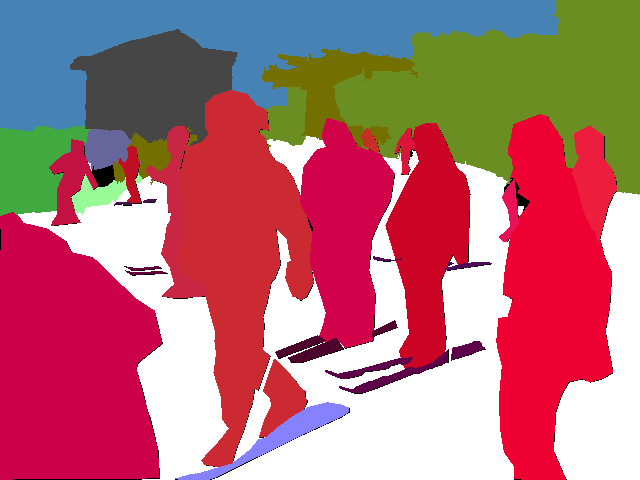}
        \caption{Ground truth}
        \label{fig:contour_gt}
    \end{subfigure} 
    \medskip
    \begin{subfigure}{0.24\textwidth}
        \includegraphics[width=\linewidth]{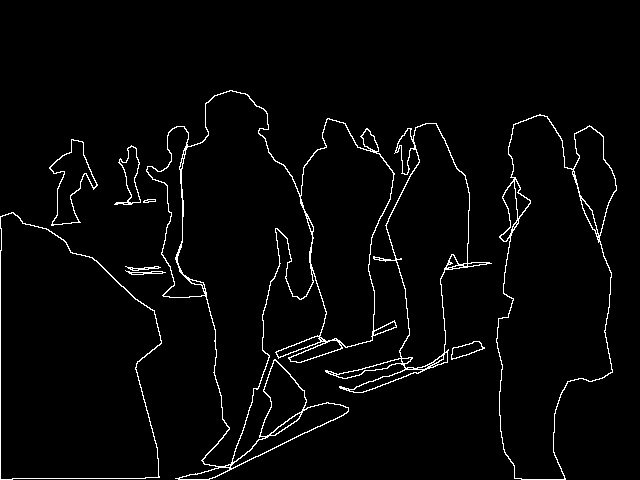}
        \caption{Silhouette feature of things}
        \label{fig:contour_thing}
    \end{subfigure} \hfil
    \begin{subfigure}{0.24\textwidth}
        \includegraphics[width=\linewidth]{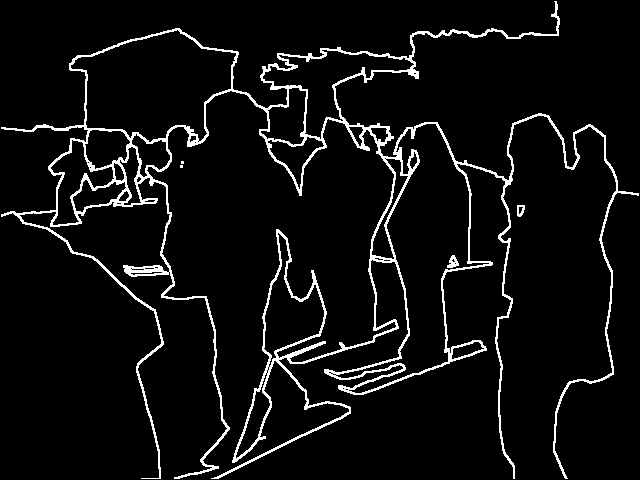}
        \caption{Silhouette feature of stuff}
        \label{fig:contour_stuff}
    \end{subfigure} 
\caption{The example of silhouette feature}
\label{fig:contour_example}
\end{figure}

Inspired by PolarMask\cite{xie2020polarmask} and Deep snake\cite{peng2020deepsnake} which formulate the instance segmentation problem as predicting contour, the silhouette can be viewed as a critical feature in the segmentation task. 
To enrich the prediction result, we proposed a silhouette-based enhancement feature. The example is shown in Fig. \ref{fig:contour_example}.

\label{sec:silhouette}

During inference time, getting the silhouette feature becomes difficult since we do not have the ground truth of the image. Therefore, it is necessary to refer to silhouette features and learn silhouette-based enhancement features during the training process, which can be viewed as binary classification on every pixel. We use the Laplacian filter as edge detection to generate the silhouette of things and stuff from the ground truth of panoptic segmentation. As discussed in \cite{deng2018boundaries}, the Dice coefficient is a better choice for predicting sharp contours. The silhouette loss is defined as follow:
\begin{equation}
\label{eq:silhouette_score}
    Score = \frac{2\sum_{i}^{H\times W} p_i g_i+\epsilon}{\sum_{i}^{H\times W} p_i^2+\sum_{i}^{H\times W} g_i^2+\epsilon}
\end{equation}
\begin{equation}
    \mathcal{L}_{silhouette}= 1-Score
\end{equation}
In the equation above, $Score$ stands for silhouette score, and the higher value means the higher similarity between the two silhouettes. $p\in H\times W$ is the silhouette-based enhancement features predicted from the model, and $g\in H\times W$ denotes the silhouette feature generated from the ground truth with the Laplacian filter. $\epsilon$ is a Laplace smooth that can be utilized to prevent division by zero and also can be used to avoid overfitting.

\subsection{Stuff Head}
\begin{figure}[!b]
    \centering
    \includegraphics[width=\linewidth]{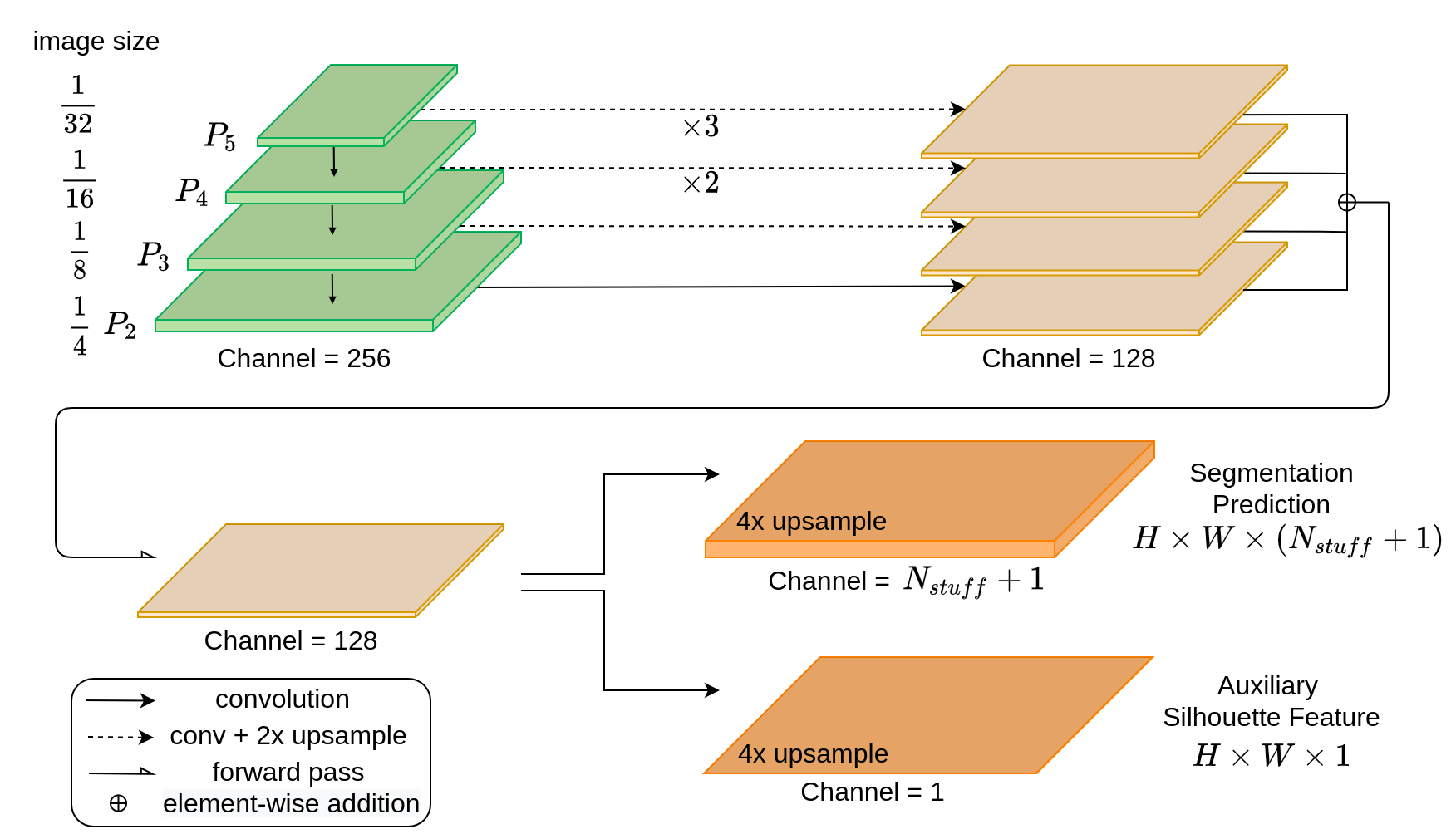}
    \caption{Stuff head}
    \label{fig:stuff_head}
\end{figure}

We use $P_2\sim P_5$ with element-wise addition to make stuff segmentation prediction. In this branch, the objects that belong to thing classes are viewed as a particular "other" class. 
Hence, the output of stuff segmentation prediction is $S\in \mathbb{R}^{H\times W\times (N_{stuff}+1)}$, where $N_{stuff}$ stands for the class number of stuff. As shown in the bottom part of Fig. \ref{fig:stuff_head}, apart from the stuff segmentation prediction, we have an auxiliary silhouette feature prediction. The output of it is $E\in \mathbb{R}^{H\times W\times 1}$. This auxiliary branch can help the stuff head feature map contain the silhouette enhancement feature. We do not need this auxiliary branch during inference, so it does not make extra time latency. We use the cross-entropy loss for our segmentation task and silhouette loss for our auxiliary branch.

\subsection{Box Head}
\begin{figure}[!b]
    \centering
    \includegraphics[width=\linewidth]{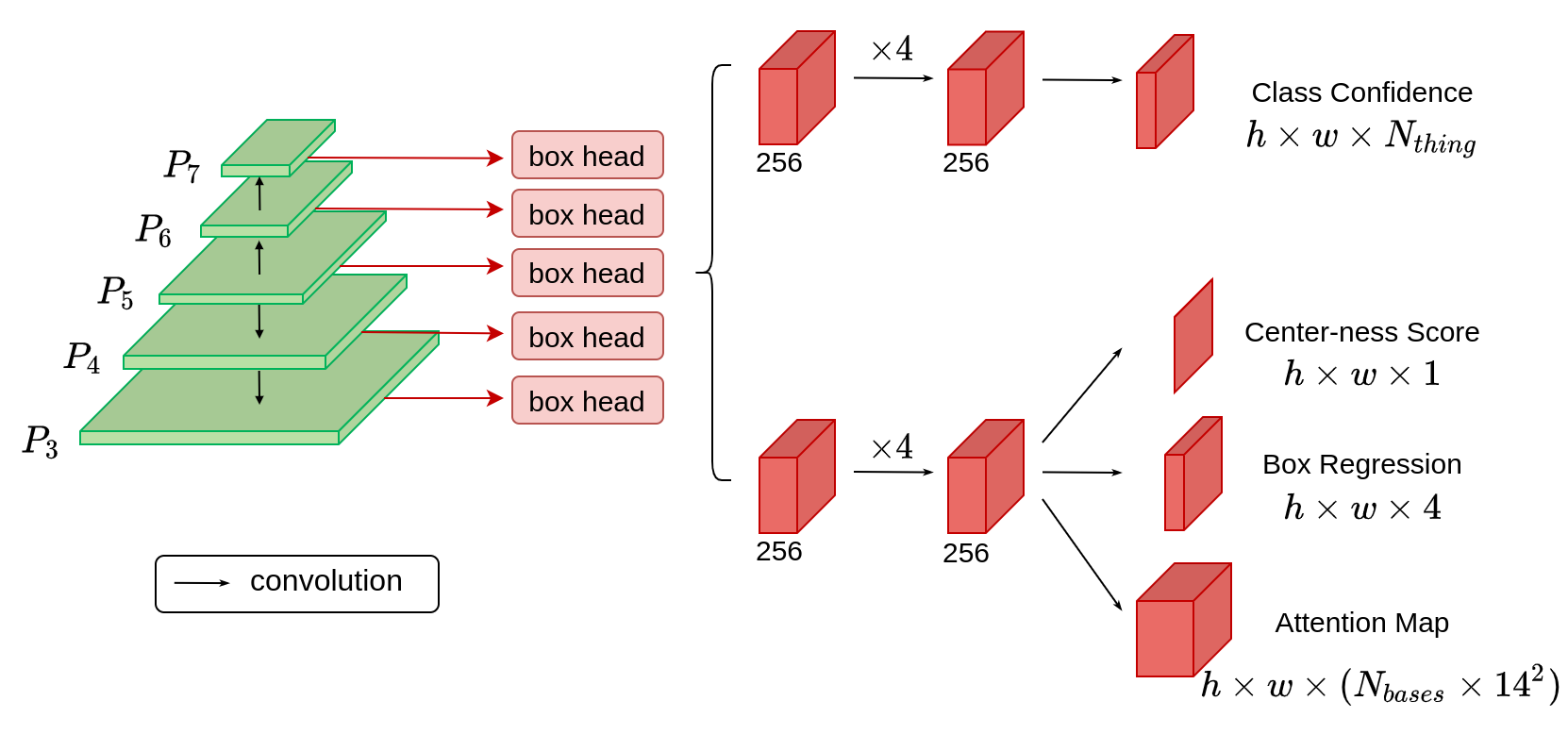}
    \caption{Box head}
    \label{fig:box_head}
\end{figure}
Inspired by the efficient instance segmentation framework BlendMask\cite{chen2020blendmask}, our instance segmentation branch predicts the mask attention maps parallel with the box detection and combines them with the bases of the image. Following BlendMask, we use FCOS\cite{tian2020fcos} as our box detection head. Here, we use $P_3\sim P_7$ from the feature pyramid network as the input. Each layer uses the same box head architecture separately to make predictions. 
As shown in Fig. \ref{fig:box_head}, there include four predictions in this head: classification confidence score, center-ness score, box position regression, and corresponding attention map for each bounding box. 
The details of them are omitted here and can be seen in BlendMask\cite{chen2020blendmask}.

\subsection{Basis Head}

\begin{figure}[t]
    \centering
    \includegraphics[width=\linewidth]{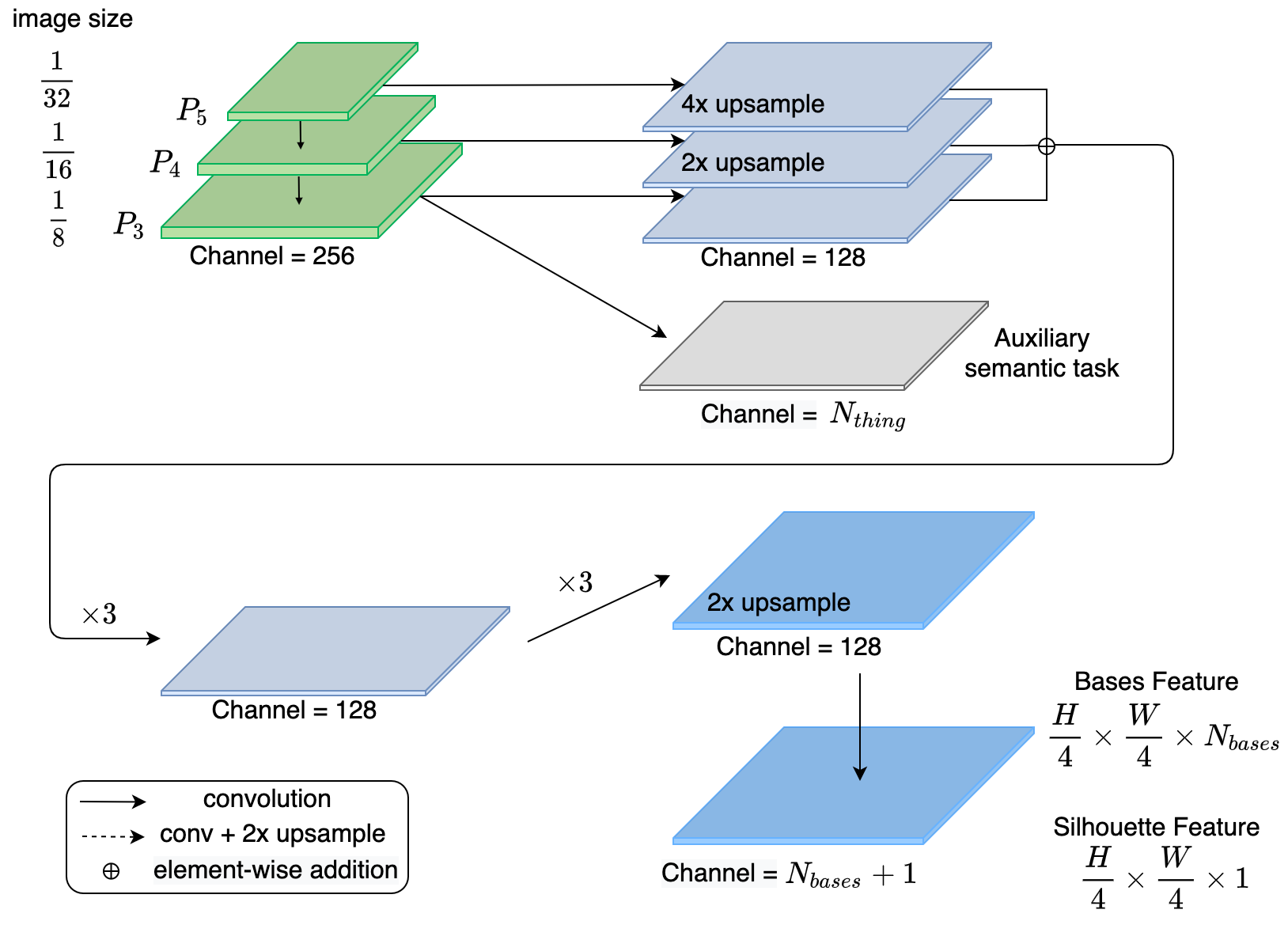}
    \caption{Basis head}
    \label{fig:basis_head}
\end{figure}

Instead of generating an instance mask for each foreground object separately, we first predict a roughly mask for the entire image. Those masks are the most critical feature in the entire image. Following the design of the stuff head, we use $P_3\sim P_5$ as our input and element-wise addition is used to combine different feature maps. 

In this branch, we also want to make sure the bases can include silhouette-based enhancement features. Hence, an additional silhouette basis has been proposed here to predict the contour of the entire image. 
On the other hand, to make sure the pyramid feature includes the instance-related feature. We add an auxiliary convolution layer with the input of feature $P_3$ and predict the semantic information for the thing classes.



\subsection{Mask Head}
Mask head aims to produce a mask for each instance detected by the box head in the image. There are two input sources for the head, including the attention map predicted by the box head and the image bases. We first do post-processing with non-maximum suppression(NMS) on the box head according to the FCOS\cite{tian2020fcos} score. After pruning, only $N_{ins}$ boxes and their corresponding attention maps will remain. 

\begin{figure}[b]
    \centering
    \includegraphics[width=\linewidth]{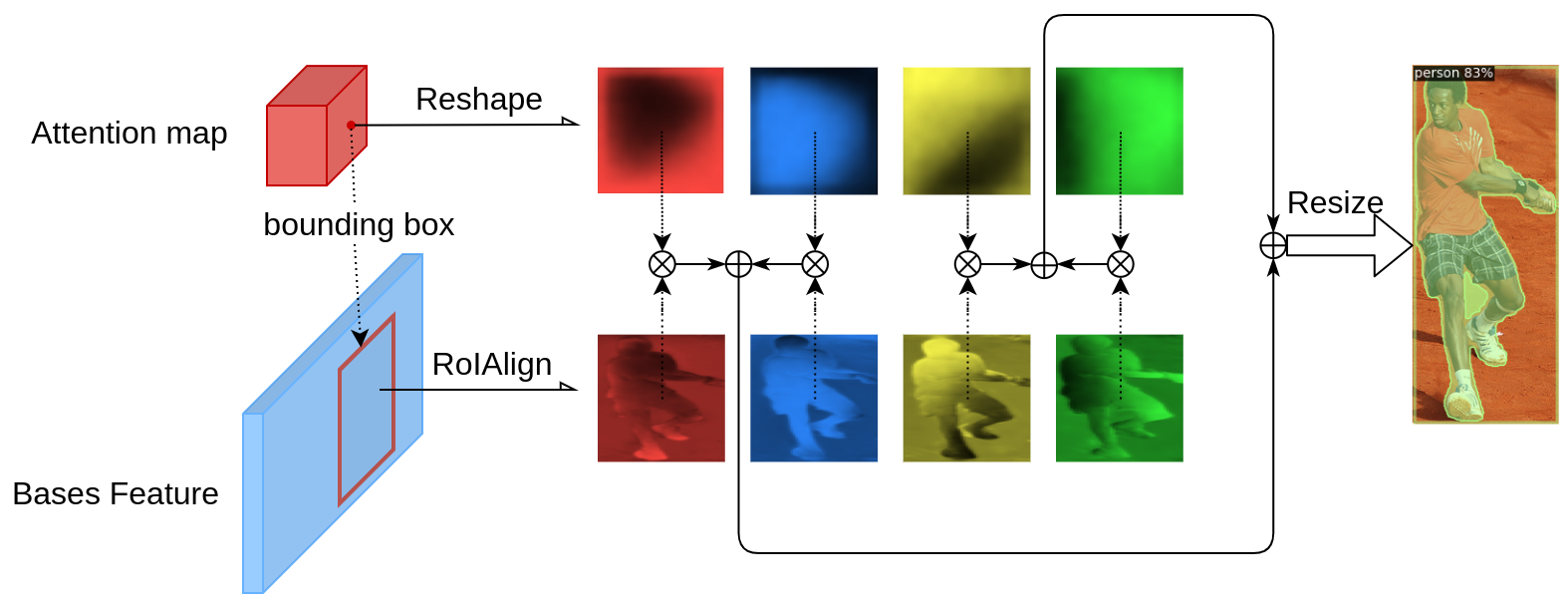}
    \caption{Schematic diagram of mask head}
    \label{fig:mask_head}
\end{figure}

Recall that our box head predicts a bounding box and attention maps for each pixel. They both use a 1-D vector to represent each pixel's prediction. First, we reshape each attention map along the channel into $N_{bases}$ 2-D image, following an up-sample operation. Then using a softmax function to normalize each attention map and get the attention score maps $S\in \mathbb{R}^{56\times56}$. Those operations are shown in the upper part of Fig. \ref{fig:mask_head}. Next, we use ROIAlign\cite{he2017maskrcnn} to crop bases feature with bounding box. Hence, we get the bases feature within the instance $F\in \mathbb{R}^{56\times56}$, which is shown in the bottom part of Fig. \ref{fig:mask_head}. 

After getting attention score maps $S$ and instance-related bases feature $F$, the instance's mask logits $M\in \mathbb{R}^{56\times56}$ can be produced by element-wise products between basis feature and corresponding attention maps, then do the element-wise addition. Finally, we can use the sigmoid function and reshape to the size same as bounding box prediction to get the instance mask prediction. The entire process is shown in Fig. \ref{fig:mask_head}. 
\begin{equation}
    M_i = \sum_{k=1}^{N_{bases}}{S_i^k\otimes F_i^k},\, \textrm{where}\; i=1\dotso N_{ins}
\end{equation}


To ensure our mask prediction in the instance segmentation task can contain silhouette features, we add an auxiliary task here. It uses the mask logits $M$ following two more convolution layers to predict the silhouette feature.

\subsection{Confidence Head}

MS R-CNN\cite{huang2019maskscore} has shown the effectiveness of using the predicted IoU score. Furthermore, since the ground truth mask of instance segmentation is contained the entire object, including the occlusion region. Hence, the IoU between the predicted mask for panoptic segmentation and the ground truth of instance segmentation will become lower when the object is occluded. Therefore, it can become a good solution in panoptic segmentation. We further extend it to the silhouette score, which we already mentioned in Eq. \ref{eq:silhouette_score}. 

\begin{figure}
    \centering
    \includegraphics[width=\linewidth]{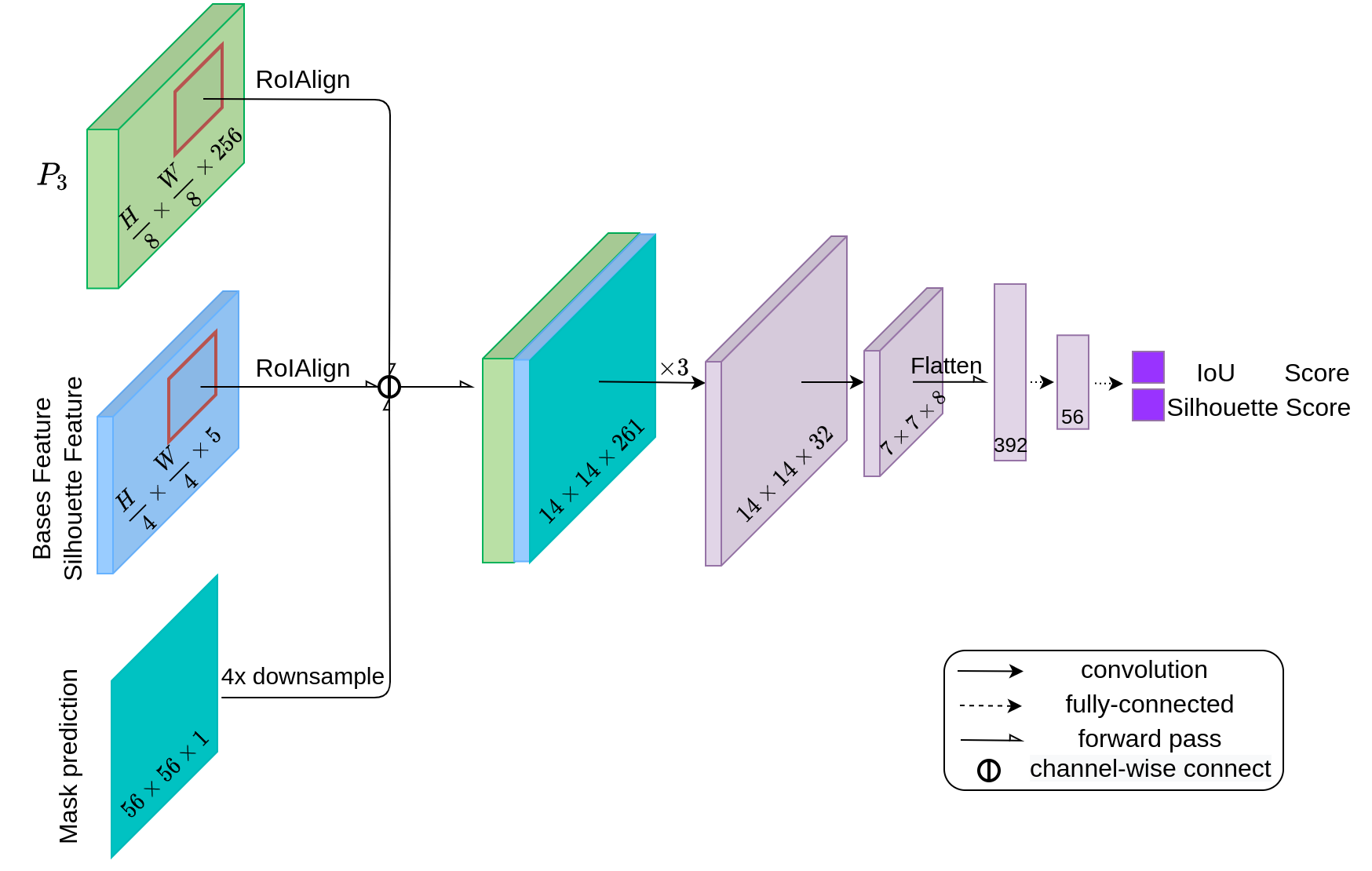}
    \caption{Confidence Head}
    \label{fig:confidence_head}
\end{figure}

As shown in Fig. \ref{fig:confidence_head}, we use three resources to produce our confidence score, namely, $P_3$ from the feature pyramid network, bases features with silhouette features, and mask prediction. First and second, we use RoIAlign\cite{he2017maskrcnn} with the instance's corresponding bounding box to crop the RoI feature from $P_3$ and bases features. Third, the mask logits $M$ from the mask head. After channel-wise connection, we use four convolution layers following two fully-connect layers to predict the IoU score and silhouette score. MSE loss is adopted as our loss function in the confidence head.

The IoU score affects globally, while the silhouette score only represents the quality of the mask. Hence, we formulate our confidence score during inference time as follows, where $\alpha$ is a hyper-parameter that balances class and IoU scores. Moreover, the silhouette score will be considered only whenever two masks possibly belong to the same instance. 
\begin{equation}
    \text{Mask Score}=\alpha \times \text{FCOS\ score}+(1-\alpha) \times \text{IoU score}
\end{equation}


\section{Experienments}
\subsection{Experimental Setup}
\subsubsection{Implementation Details}
We implement our method based on Pytorch with batch size 6 on a single GPU RTX 3090. We freeze the batch normalization layers in the backbone and use group normalization layers\cite{wu2018group} within the different heads. The entire model is being trained in an end-to-end fashion for 600k steps. The optimizer adopted SGD, and the initial learning rate is set to 0.01 with a constant warm-up of 1k iterations. Weight decay and momentum are set as 0.0001 and 0.9, respectively. The learning rate is reduced by a factor of 10 at iteration 120k, 420k, and 500k. The image is randomly flip and rescale the shorter edge from 640 to 800 pixels.
\subsubsection{Metrics.}
We adopt the evaluation metric called panoptic quality (PQ), which is introduced by \cite{kirillov2019panoptic}. Panoptic quality is defined as:
\begin{equation}
    PQ = \underbrace{\frac{\sum_{(p,g)\in \text{TP}}\text{IoU}(p,g)}{|\text{TP}|} }_{\text{SQ}}  
    \underbrace{\frac{|\text{TP}|}{|\text{TP}| + \frac{1}{2}|\text{FP}| + \frac{1}{2}|\text{FN}|}}_{\text{RQ}},
\end{equation}
\subsection{Ablation Study}
\begin{table*}[!t]
\centering
\caption{Ablation study on CityScapes panoptic \emph{val} set.}
\begin{tabular}{c c c c c | c c c c c}
\hline
\makecell{Mask\\Loss} & \makecell{Mask\\Score} & \makecell{\small Silhouette\\Feature} & \makecell{Mask\\Aux} & \makecell{\small Silhouette\\Stuff} & PQ & $\text{PQ}^{\text{Th}}$ & $\text{PQ}^{\text{St}}$ & SQ & RQ  \\ \hline \hline
 -          &      -     &     -      &     -      &     -      & 57.88 & 55.18 & 59.84 & 79.99 & 70.74  \\ \hline %
 \checkmark &            &            &            &            & 58.25 & 55.45 & 60.29 & 80.04 & 71.12 \\ \hline
 \checkmark & \checkmark &            &            &            & 58.44 & \textbf{55.90} & 60.30 & 80.06 & 71.35 \\ \hline
 \checkmark &            & \checkmark &            &            & 58.91 & 55.76 & 61.20 & 80.27 & 71.85 \\ \hline
 \checkmark & \checkmark & \checkmark &            &            & 59.10 & 55.77 & 61.53 & 80.01 & 72.36 \\ \hline
 \checkmark &            & \checkmark & \checkmark &            & 59.00 & 55.50 & 61.54 & 80.43 & 71.87 \\ \hline %
 \checkmark & \checkmark & \checkmark & \checkmark &            & 59.15 & 55.73 & 61.63 & 80.63 & 71.86 \\ \hline %
 \hline 
 
 \checkmark &            &            &            & \checkmark & 59.03 & 55.27 & 61.76 & 80.51 & 71.95 \\ \hline %
 \checkmark & \checkmark &            &            & \checkmark & 59.07 & 55.33 & 61.78 & \textbf{80.64} & 71.90 \\ \hline %
 \checkmark &            & \checkmark &            & \checkmark & 59.51 & 55.48 & 62.44 & 80.52 & 72.56 \\ \hline %
 \checkmark & \checkmark & \checkmark &            & \checkmark & 59.60 & 55.70 & 62.44 & 80.48 & 72.70 \\ \hline %
 \checkmark &            & \checkmark & \checkmark & \checkmark & 59.83 & 55.65 & 62.87 & 80.47 & 72.95 \\ \hline %
 \checkmark & \checkmark & \checkmark & \checkmark & \checkmark & \textbf{59.95} & 55.87 & \textbf{62.91} & 80.50 & \textbf{73.06} \\ \hline %

\end{tabular}
\label{table:ablation-city-panoptic}
\end{table*}
To verify the performance of our proposed, we conduct the experiments with different settings in Table \ref{table:ablation-city-panoptic} on the CityScapes \emph{val} set with single GPU GTX 1080ti. We adopted BlendMask\cite{chen2020blendmask} with stuff branch as our baseline model. 

\subsubsection{Mask Loss}
In other instance segmentation works, they often use cross-entropy as the sole loss function. Different from them, we implemented IoU loss and silhouette loss additionally to guide the instance mask prediction to have a higher quality in area and shape. Since the instance branch and the stuff branch have a shared backbone, we can find out that the higher instance mask quality not only can improve PQ$^{\text{Th}}$ but also increase PQ$^{\text{St}}$. 

\subsubsection{Mask Score}
With the mask score module, we are using the confidence score produced by the confidence head. In other words, apart from the class confidence score, we also pay attention to the IoU score and silhouette score. The $\alpha$ is set to 0.1 in the experiments. In our experiments, we can see that RQ can increase with a small margin in most situations. Hence, it can be explained as using the confidence head to get more precise detection since the score is more representative of the things.

\subsubsection{Silhouette Feature}

We add an extra silhouette-based enhancement feature in the basis head. It can help the model learn the silhouette feature for the entire image in the earlier step. As shown in the table \ref{table:ablation-city-panoptic}, after adding the module can markedly improve PQ$^{\text{Th}}$. This represents that the silhouette-based enhancement feature has a large impact on the instance mask prediction quality.

\subsubsection{Mask Auxiliary}
We use an auxiliary task in our mask head to guarantee the mask prediction contains a silhouette-based feature. Since it does not need computation during the inference time, it will not be a burden. Moreover, we can find that adding this module into our model can improve significantly, especially in semantic quality(SQ). 

\subsubsection{Silhouette Stuff}
We also guide the stuff branch to get a better mask prediction by silhouette feature. Compare with the upper part and the lower part of the table, we can find out that an extra auxiliary silhouette feature prediction can significantly improve PQ$^{\text{St}}$. 

\subsection{Analysis of Confidence Head}
\begin{figure}[t]
\minipage{0.24\textwidth}
\centering
  \includegraphics[width=\linewidth]{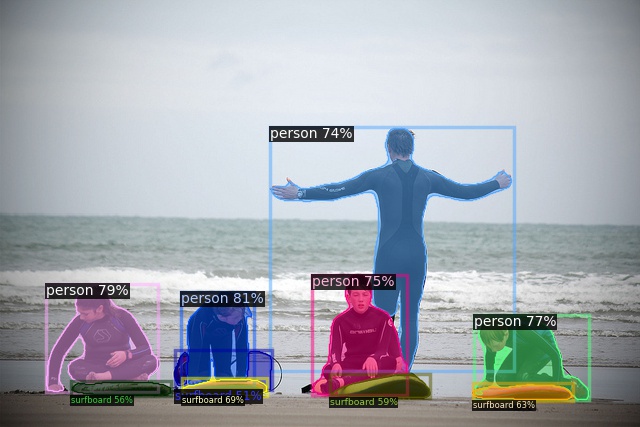}
  Class Score
\endminipage\hfill
\minipage{0.24\textwidth}
\centering
  \includegraphics[width=\linewidth]{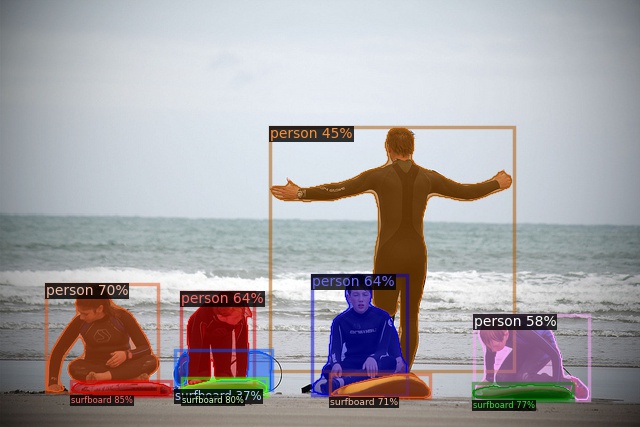}
  Confidence Score
\endminipage\hfill
\caption{Visualization results of IoU score prediction on COCO panoptic \emph{val} set. The best view zoomed in on a color screen.}
\label{fig:iou_result}
\end{figure}

As shown in Fig. \ref{fig:iou_result}, we give the prediction result of the confidence score from the confidence head to verify the effect. In the example, we can see that surfboards have higher confidence scores than the person. This means that surfboards have a higher chance of getting the entire mask prediction without any occlusion. In contrast, the person standing and back to the picture has a lower confidence score since people in front obscure his legs. Hence, the proposed confidence score can be more representative of instance reality.

On the other hand, as shown in Table \ref{table:alpha_ablation}, we do the experiments on the confidence head hyperparameters $\alpha$, which balances class and IoU scores, with our ablation study setting. It shows that just a tiny weight of IoU score can help to improve the performance. We set the $\alpha$ equal to 0.8 in all other experiments if not specify.

\begin{table}[!t]
\centering
\caption{Results with different $\alpha$ on CityScapes panoptic \emph{val} set }
\begin{tabular}{c | c c c c c}
\hline
$\alpha$ & PQ & $\text{PQ}^{\text{Th}}$ & $\text{PQ}^{\text{St}}$ & SQ & RQ \\ \hline \hline
0.1 & 56.29 & 47.18 & 62.91 & 79.85 & 69.01 \\ \hline 
0.2 & 57.11 & 49.14 & 62.91 & 79.92 & 70.03 \\ \hline 
0.3 & 57.78 & 50.71 & \textbf{62.93} & 80.02 & 70.81 \\ \hline 
0.4 & 57.98 & 51.22 & 62.91 & 80.08 & 71.02 \\ \hline 
0.5 & 59.07 & 53.79 & 62.90 & 80.26 & 72.20 \\ \hline 
0.6 & 59.61 & 55.10 & 62.89 & 80.44 & 72.72 \\ \hline 
0.7 & 59.71 & 55.35 & 62.88 & 80.45 & 72.83 \\ \hline 
0.8 & \textbf{59.95} & \textbf{55.87} & 62.91 & 80.50 & \textbf{73.06} \\ \hline 
0.9 & 59.77 & 55.48 & 62.89 & 80.55 & 72.81 \\ \hline 
1.0 & 59.55 & 54.94 & 62.90 & \textbf{80.62} & 72.49 \\ \hline 
\end{tabular}
\label{table:alpha_ablation}
\end{table}

\subsection{Analysis of Silhouette Feature}
\begin{figure}[t]
\small
\minipage{0.16\textwidth}
\centering
  \includegraphics[width=\linewidth, height=0.8\linewidth]{figures/Experiments/image/000000013291.jpg}
\endminipage\hfill
\minipage{0.16\textwidth}
\centering
  \includegraphics[width=\linewidth, height=0.8\linewidth]{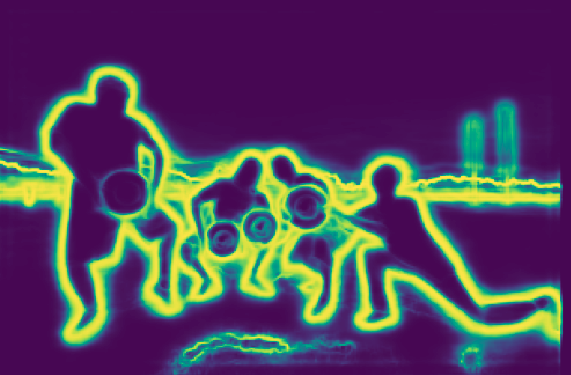}
\endminipage\hfill
\minipage{0.16\textwidth}%
\centering
  \includegraphics[width=\linewidth, height=0.8\linewidth]{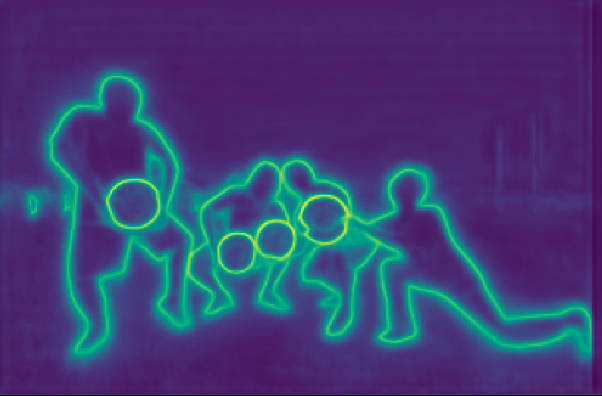}
\endminipage \hfill

\minipage{0.16\textwidth}
\centering
  \includegraphics[width=\linewidth, height=0.8\linewidth]{figures/Experiments/image/000000436738.jpg}
  Image
\endminipage\hfill
\minipage{0.16\textwidth}
\centering
  \includegraphics[width=\linewidth, height=0.8\linewidth]{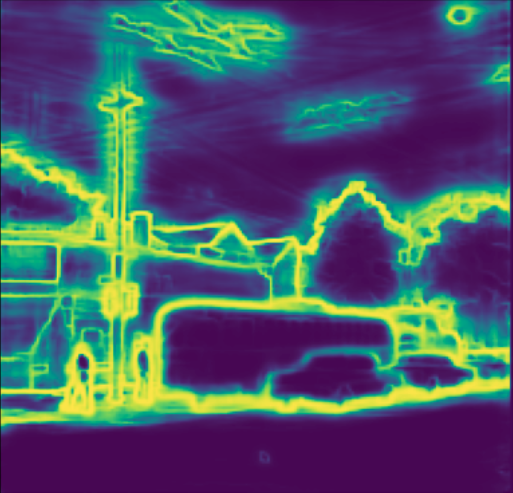}
  Stuff
\endminipage\hfill
\minipage{0.16\textwidth}%
\centering
  \includegraphics[width=\linewidth, height=0.8\linewidth]{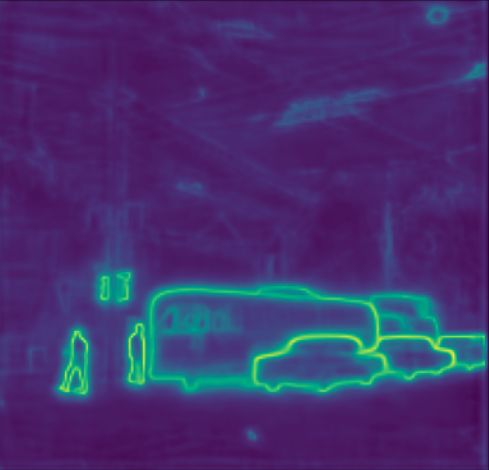}
  Thing
\endminipage \hfill

\caption{Visualization results of silhouette feature on COCO panoptic \emph{val} set}
\label{fig:silhouette_result}
\end{figure}
We further make a visualization of what does the model has learned. As shown in Fig. \ref{fig:silhouette_result}, we separate the silhouette feature of stuff and thing. The silhouette for the things can pay more attention to the foreground object but omits the background information. However, the stuff-related silhouette features cannot distinguish different instances well. In contrast, they have a higher reaction on every junction, regardless of foreground or background.

\subsection{Qualitative Results}
\begin{figure}[!t]
\minipage{0.24\textwidth}%
\centering
  \includegraphics[width=\linewidth]{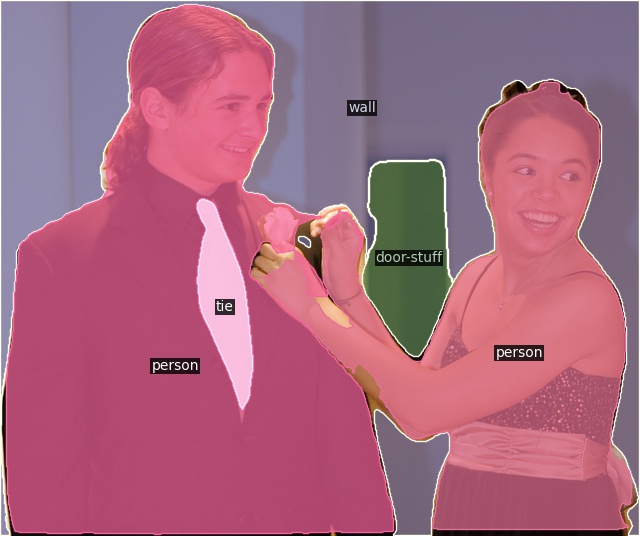}
\endminipage  \hfill
\minipage{0.24\textwidth}%
\centering
  \includegraphics[width=\linewidth]{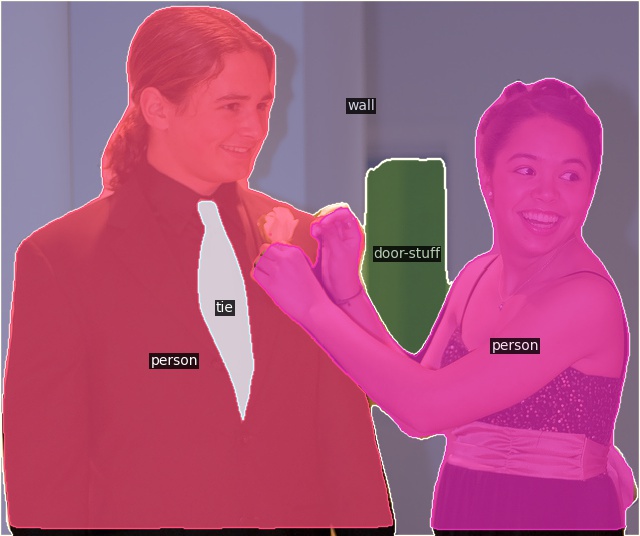}
\endminipage  \hfill

\minipage{0.24\textwidth}%
\centering
  \includegraphics[width=\linewidth]{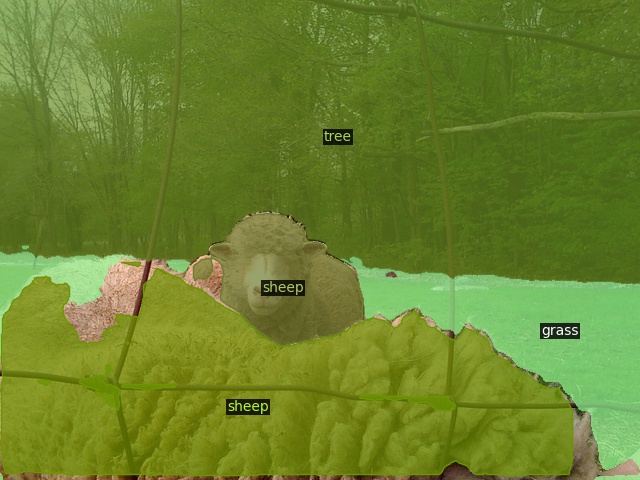}
\endminipage  \hfill
\minipage{0.24\textwidth}%
\centering
  \includegraphics[width=\linewidth]{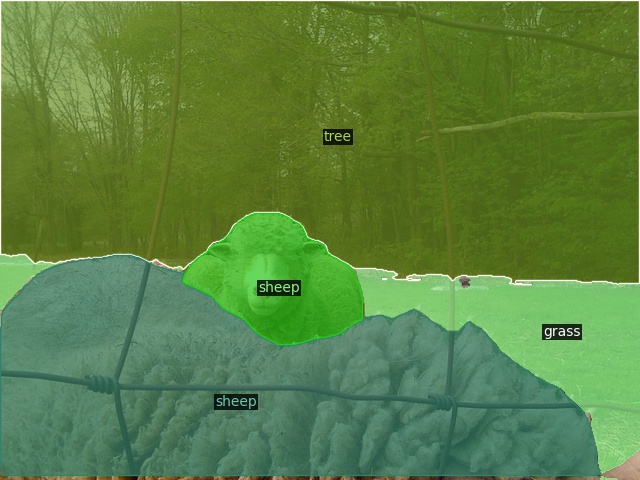}
\endminipage  \hfill

\minipage{0.24\textwidth}%
\centering
  \includegraphics[width=\linewidth]{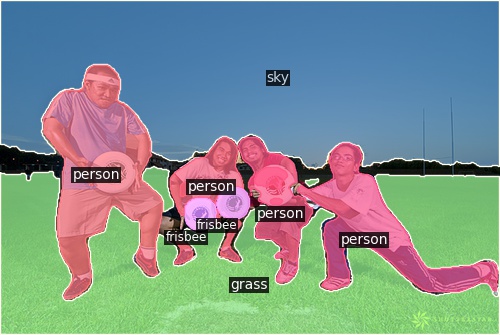}
  Baseline
\endminipage  \hfill
\minipage{0.24\textwidth}%
\centering
  \includegraphics[width=\linewidth]{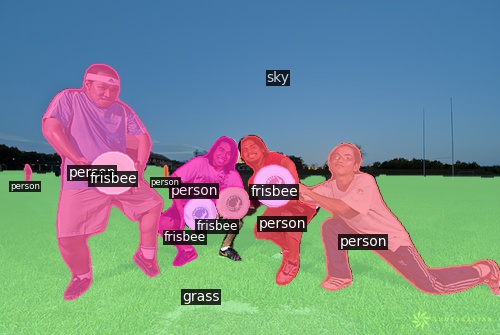}
  Ours
\endminipage  \hfill
\caption{Qualitative results compare to the baseline model}
\label{fig:qualitative_result}
\end{figure}
We give some qualitative results of SE-PSNet, as presented in Fig. \ref{fig:qualitative_result}. These results show that our model can have a better quality mask, especially on the silhouette of the instance. For instance, the woman's hands can be segmented well in the first row, and the sheep in front of the second row have the same situation. Moreover, the first row shows the importance of the proposed confidence score when the big object overlaps with a small instance, i.e., man and tie.

\subsection{Quantitative Results}
\begin{table*}[ht]
\centering
\caption{Panoptic segmentation results on COCO panoptic dataset }
\begin{tabular}{| c || c | c c c | c || c | c c c c c|}
\hline
&\multicolumn{5}{c||}{\text{Val} } & \multicolumn{6}{c|}{\text{Test-dev} } \\ \hline
Method & Backbone & PQ &  $\text{PQ}^{\text{Th}}$ & $\text{PQ}^{\text{St}}$  & time (ms) & Backbone & PQ & $\text{PQ}^{\text{Th}}$ & $\text{PQ}^{\text{St}}$ & SQ & RQ \\ \hline 
 \multicolumn{12}{|c|}{\text{Proposed-free} } \\ \hline

DeeperLab\cite{yang2019deeperlab} & Xcep-71 & 33.8 & - & - & 94 & Xcep-71 & 34.3 & 37.5 & 29.6 & 77.1 & 43.1 \\
Panoptic-DeepLab\cite{Cheng2020PanopticDeepLabAS} & Xcep-71 & 39.7 & 43.9 & 33.2 & 132 &  Xcep-71  & 41.4 & 45.1 & 35.9 & - & - \\
SSAP\cite{gao2019ssap} & R-101 & 36.5 & - & - & - &  R-101  & 36.9 & 40.1 & 32.0 & 80.7 & 44.8 \\ 
Axial-DeepLab\cite{wang2020axial} & Axial-R-L & 43.4 & 48.5 & 35.6 & - & Axial-R-L & 43.6 & 48.9 & 35.6 & - & -  \\ 
Panoptic FCN\cite{li2021panopticfcn} & R-50 & 43.6 & 50.0 & 35.6 & 80 &  R-101  & 45.5 & 51.4 & 36.4 & - & -  \\ \hline

\multicolumn{12}{|c|}{\text{Proposed-based} } \\ \hline

JSIS\cite{de2018jsis}  & R-50 &  26.9 & 29.3 & 23.3 & - &  R-50 & 27.2 & 29.6 & 23.4 & 71.9 & 35.9 \\
TASCNet\cite{li2018tascnet}  & - & - & - & - & - &  R-101  & 40.7 & 47.0 & 31.0 & 78.5 & 50.1 \\
AUNet\cite{li2019aunet} & R-50 & 39.6 &  49.1 &  25.2  &   - &  R-101 & 45.2 & 54.4 & 32.5 & 81.0 & 56.1 \\
Panoptic-FPN\cite{kirillov2019panopticfpn}  & R-101 & 40.3 & 47.5 &  29.5  &  - &  R-101  & 40.9 & 48.3 & 29.7 & - & - \\
UPSNet\cite{Xiong2019UPSNetAU}  & R-50 & 42.5 &  48.6 &  33.4  & 167 &  R-101$^*$  & 46.6 & 53.2 & 36.7 & 80.5 & 56.9 \\
OANet\cite{liu2019oanet}  & R-101 & 40.7 & \textbf{59.9} & 26.6 & - &  R-101 & 41.3 & 50.4 & 27.7 & - & - \\
OCFusion\cite{lazarow2020ocfusion}  & R-101 & 43.0 & 51.1 & 30.7 & 156 &  X-101$^*$ & 46.7 & 54.0 & 35.7 & - & - \\
AdaptIS\cite{sofiiuk2019adaptis}  & X-101 & 42.3 & 49.2 &  31.8  &  - &  X-101  & 42.8 & 50.1 & 31.8 & - & - \\
SOGNet\cite{yang2020sognet}  & R-50 & 43.7 & 50.6 & 33.2 & 179$^\#$ &  R-101$^*$  & 47.8 & - & - & 80.7 & 57.6  \\
EPSNet\cite{chang2020epsnet}  & R-101 & 38.6 & 43.5 & 31.3 & \textbf{51} &  R-101  & 38.9 & 44.1 & 31.0 & -  & -  \\
Unifying\cite{Li2020UnifyingTA}  & R-50 & 43.4 & 48.6 & \textbf{35.5} & - &  R-101 & 47.2 & 53.5 & 37.7 & 81.1 & 57.2  \\
CondInst\cite{tian2021conditional}  & - & - & - & - & - &  R-101  & 46.1 & 54.7 & 33.2 & -  & -  \\ \hline
Ours & R-101 & \textbf{44.4}  &  51.8  &  33.2 &  75 &  X-101  &  46.4  &  53.8  &  35.3 & \textbf{81.7} & 55.4  \\ \hline
\multicolumn{12}{l}{$^*$ Using deformable convolution in the backbone.\; $^\#$ We use the code published by the author to calculate with RTX 2080ti.}
\end{tabular}
\label{table:COCO_results}
\end{table*}

\begin{table}[ht]
\centering
\caption{Instance segmentation results on COCO \emph{test-dev} }
\begin{tabular}{ c | c | c c c }
\hline
Method & Backbone & AP & $\text{AP}_{50}$ & $\text{AP}_{75}$ \\ \hline \hline
\multicolumn{5}{c}{\text{Two-stage} } \\ \hline
Mask R-CNN\cite{he2017maskrcnn} & R-101-FPN & 35.7 & 58.0 & 37.8  \\
MS R-CNN\cite{huang2019maskscore} & R-101-FPN & 38.3 & 58.8 & 41.5  \\
HTC\cite{chen2019hybrid} & R-101-FPN & 39.7 & 61.8 & 43.1  \\
BMask R-CNN\cite{cheng2020boundary} & R-101-FPN & 37.7 & 59.3 & 40.6  \\
BCNet\cite{ke2021bcnet} & R-101-FPN & 39.8 & 61.5 & 43.1  \\ 

\hline

\multicolumn{5}{c}{\text{One-stage} } \\ \hline
YOLACT\cite{bolya2019yolact} & R-101-FPN & 31.2 & 50.6 & 32.8  \\
TensorMask\cite{chen2019tensormask} & R-101-FPN & 37.1 & 59.3 & 39.4  \\
PolarMask\cite{xie2020polarmask} & R-101-FPN & 32.1 & 53.7 & 33.1  \\
CenterMask (Wang)\cite{wang2020centermask} & Hourglass-104 & 34.5 & 56.1 & 36.3  \\
CenterMask (Lee)\cite{lee2020centermask} & R-101-FPN & 38.3 & - & -  \\
SOLO\cite{wang2020solo} & R-101-FPN & 37.8 & 59.5 & 40.4 \\
BlendMask\cite{chen2020blendmask} & R-101-FPN & 38.4 & 60.7 & 41.3  \\
SOLOv2\cite{wang2020solov2} & R-101-FPN & 39.7 & 60.7 & 42.9  \\
CondInst\cite{tian2021conditional} & R-101-FPN & 39.1 & 60.8 & 41.9  \\ 
\hline
Ours &  R-101-FPN  &  39.3  &  60.6  &  42.3  \\ 
Ours &  X-101-FPN  &  \textbf{41.4}  &  \textbf{63.1}  &  \textbf{44.7}  \\ \hline
\end{tabular}
\label{table:COCO_in_test_results}
\end{table}

We compare our network on COCO panoptic dataset with panoptic quality and inference speed. Specifically, we test with a single-scale $800\times1333$ image.
As shown in Table \ref{table:COCO_results}, compared with recent approaches, SE-PSNet has a fast inference time since our proposed module does not need extra effort during inference time. In addition, we get the competitive result on the overall performance. Without performing horizontal flipping and multi-scale input images for testing, we outperform the others on the semantic quality, which is the main issue we want to improve.

We further do additional experiments on instance segmentation to show its efficiency of mask quality. As shown in Table \ref{table:COCO_in_test_results}, apart from AP, we get a good performance in AP$_{75}$. It represents that when the standard becomes stricter, we can still have a high-quality performance since we pay attention to the mask quality and silhouette feature during training. 

\section{Conclusion}
In this work, we propose a Silhouette-based Enhancement Feature for Panoptic Segmentation Network, which tackles the irregular mask prediction near the boundary. 
The proposed silhouette feature aims to focus on the contour of masks, and the new mask score can be more representative of mask quality. Furthermore, the auxiliary task used to guide the prediction brings improvement without causing overhead during inference. 
We do experiments on the COCO dataset and CityScapes dataset. Those, including panoptic segmentation and instance segmentation, show that we achieve competitive performance with a much faster inference time. Also, SE-PSNet is able to predict a high-quality mask on the boundary of instance which help us distinguish different instance easier. In the future, we will further extend this work to autonomous driving, helping the computer detect the boundary of different instances better.

\bibliographystyle{ieeetr}
\bibliography{references}

\begin{thebibliography}{10}

\bibitem{he2017maskrcnn}
K.~He, G.~Gkioxari, P.~Doll{\'a}r, and R.~Girshick, ``Mask r-cnn,'' in {\em
  Proceedings of the IEEE international conference on computer vision},
  pp.~2961--2969, 2017.

\bibitem{li2018tascnet}
J.~Li, A.~Raventos, A.~Bhargava, T.~Tagawa, and A.~Gaidon, ``Learning to fuse
  things and stuff,'' {\em arXiv preprint arXiv:1812.01192}, 2018.

\bibitem{li2019aunet}
Y.~Li, X.~Chen, Z.~Zhu, L.~Xie, G.~Huang, D.~Du, and X.~Wang,
  ``Attention-guided unified network for panoptic segmentation,'' in {\em
  Proceedings of the IEEE/CVF Conference on Computer Vision and Pattern
  Recognition}, pp.~7026--7035, 2019.

\bibitem{kirillov2019panopticfpn}
A.~Kirillov, R.~Girshick, K.~He, and P.~Doll{\'a}r, ``Panoptic feature pyramid
  networks,'' in {\em Proceedings of the IEEE/CVF Conference on Computer Vision
  and Pattern Recognition}, pp.~6399--6408, 2019.

\bibitem{lin2017fpnnet}
T.-Y. Lin, P.~Doll{\'a}r, R.~Girshick, K.~He, B.~Hariharan, and S.~Belongie,
  ``Feature pyramid networks for object detection,'' in {\em Proceedings of the
  IEEE conference on computer vision and pattern recognition}, pp.~2117--2125,
  2017.

\bibitem{Xiong2019UPSNetAU}
Y.~Xiong, R.~Liao, H.~Zhao, R.~Hu, M.~Bai, E.~Yumer, and R.~Urtasun, ``Upsnet:
  A unified panoptic segmentation network,'' {\em 2019 IEEE/CVF Conference on
  Computer Vision and Pattern Recognition (CVPR)}, pp.~8810--8818, 2019.

\bibitem{liu2019oanet}
H.~Liu, C.~Peng, C.~Yu, J.~Wang, X.~Liu, G.~Yu, and W.~Jiang, ``An end-to-end
  network for panoptic segmentation,'' in {\em Proceedings of the IEEE/CVF
  Conference on Computer Vision and Pattern Recognition}, pp.~6172--6181, 2019.

\bibitem{lazarow2020ocfusion}
J.~Lazarow, K.~Lee, K.~Shi, and Z.~Tu, ``Learning instance occlusion for
  panoptic segmentation,'' in {\em Proceedings of the IEEE/CVF Conference on
  Computer Vision and Pattern Recognition}, pp.~10720--10729, 2020.

\bibitem{sofiiuk2019adaptis}
K.~Sofiiuk, O.~Barinova, and A.~Konushin, ``Adaptis: Adaptive instance
  selection network,'' in {\em Proceedings of the IEEE/CVF International
  Conference on Computer Vision}, pp.~7355--7363, 2019.

\bibitem{yang2020sognet}
Y.~Yang, H.~Li, X.~Li, Q.~Zhao, J.~Wu, and Z.~Lin, ``Sognet: Scene overlap
  graph network for panoptic segmentation,'' in {\em Proceedings of the AAAI
  Conference on Artificial Intelligence}, vol.~34, pp.~12637--12644, 2020.

\bibitem{chang2020epsnet}
C.-Y. Chang, S.-E. Chang, P.-Y. Hsiao, and L.-C. Fu, ``Epsnet: Efficient
  panoptic segmentation network with cross-layer attention fusion,'' in {\em
  Proceedings of the Asian Conference on Computer Vision}, 2020.

\bibitem{Li2020UnifyingTA}
Q.~Li, X.~Qi, and P.~Torr, ``Unifying training and inference for panoptic
  segmentation,'' {\em 2020 IEEE/CVF Conference on Computer Vision and Pattern
  Recognition (CVPR)}, pp.~13317--13325, 2020.

\bibitem{tian2021conditional}
Z.~Tian, B.~Zhang, H.~Chen, and C.~Shen, ``Instance and panoptic segmentation
  using conditional convolutions,'' {\em ArXiv}, vol.~abs/2102.03026, 2021.

\bibitem{mohan2021efficientps}
R.~Mohan and A.~Valada, ``Efficientps: Efficient panoptic segmentation,'' {\em
  International Journal of Computer Vision}, pp.~1--29, 2021.

\bibitem{yang2019deeperlab}
T.-J. Yang, M.~D. Collins, Y.~Zhu, J.-J. Hwang, T.~Liu, X.~Zhang, V.~Sze,
  G.~Papandreou, and L.-C. Chen, ``Deeperlab: Single-shot image parser,'' {\em
  arXiv preprint arXiv:1902.05093}, 2019.

\bibitem{chen2018deeplabv3+}
L.-C. Chen, Y.~Zhu, G.~Papandreou, F.~Schroff, and H.~Adam, ``Encoder-decoder
  with atrous separable convolution for semantic image segmentation,'' in {\em
  Proceedings of the European conference on computer vision (ECCV)},
  pp.~801--818, 2018.

\bibitem{Cheng2020PanopticDeepLabAS}
B.~Cheng, M.~D. Collins, Y.~Zhu, T.~Liu, T.~Huang, H.~Adam, and L.-C. Chen,
  ``Panoptic-deeplab: A simple, strong, and fast baseline for bottom-up
  panoptic segmentation,'' {\em 2020 IEEE/CVF Conference on Computer Vision and
  Pattern Recognition (CVPR)}, pp.~12475--12485, 2020.

\bibitem{gao2019ssap}
N.~Gao, Y.~Shan, Y.~Wang, X.~Zhao, Y.~Yu, M.~Yang, and K.~Huang, ``Ssap:
  Single-shot instance segmentation with affinity pyramid,'' in {\em 2019
  IEEE/CVF International Conference on Computer Vision (ICCV)}, pp.~642--651,
  2019.

\bibitem{li2021panopticfcn}
Y.~Li, H.~Zhao, X.~Qi, L.~Wang, Z.~Li, J.~Sun, and J.~Jia, ``Fully
  convolutional networks for panoptic segmentation,'' in {\em IEEE Conference
  on Computer Vision and Pattern Recognition (CVPR)}, 2021.

\bibitem{deng2018boundaries}
R.~Deng, C.~Shen, S.~Liu, H.~Wang, and X.~Liu, ``Learning to predict crisp
  boundaries,'' in {\em Proceedings of the European Conference on Computer
  Vision (ECCV)}, pp.~562--578, 2018.

\bibitem{han2021edgenet}
H.-Y. Han, Y.-C. Chen, P.-Y. Hsiao, and L.-C. Fu, ``Using channel-wise
  attention for deep cnn based real-time semantic segmentation with class-aware
  edge information,'' {\em IEEE Transactions on Intelligent Transportation
  Systems}, vol.~22, pp.~1041--1051, 2021.

\bibitem{zimmermann2019faster}
R.~S. Zimmermann and J.~N. Siems, ``Faster training of mask r-cnn by focusing
  on instance boundaries,'' {\em Computer Vision and Image Understanding},
  vol.~188, p.~102795, 2019.

\bibitem{cheng2020boundary}
T.~Cheng, X.~Wang, L.~Huang, and W.~Liu, ``Boundary-preserving mask r-cnn,'' in
  {\em European Conference on Computer Vision}, pp.~660--676, Springer, 2020.

\bibitem{peng2020deepsnake}
S.~Peng, W.~Jiang, H.~Pi, X.~Li, H.~Bao, and X.~Zhou, ``Deep snake for
  real-time instance segmentation,'' in {\em Proceedings of the IEEE/CVF
  Conference on Computer Vision and Pattern Recognition}, pp.~8533--8542, 2020.

\bibitem{cheng2021boundary}
B.~Cheng, R.~Girshick, P.~Doll{\'a}r, A.~C. Berg, and A.~Kirillov, ``Boundary
  {IoU}: Improving object-centric image segmentation evaluation,'' in {\em IEEE
  Conference on Computer Vision and Pattern Recognition (CVPR)}, 2021.

\bibitem{deng2009imagenet}
J.~Deng, W.~Dong, R.~Socher, L.-J. Li, K.~Li, and L.~Fei-Fei, ``Imagenet: A
  large-scale hierarchical image database,'' in {\em Proceedings of the IEEE
  conference on computer vision and pattern recognition}, pp.~248--255, IEEE,
  2009.

\bibitem{xie2020polarmask}
E.~Xie, P.~Sun, X.~Song, W.~Wang, X.~Liu, D.~Liang, C.~Shen, and P.~Luo,
  ``Polarmask: Single shot instance segmentation with polar representation,''
  in {\em Proceedings of the IEEE/CVF conference on computer vision and pattern
  recognition}, pp.~12193--12202, 2020.

\bibitem{chen2020blendmask}
H.~Chen, K.~Sun, Z.~Tian, C.~Shen, Y.~Huang, and Y.~Yan, ``Blendmask: Top-down
  meets bottom-up for instance segmentation,'' in {\em Proceedings of the
  IEEE/CVF conference on computer vision and pattern recognition},
  pp.~8573--8581, 2020.

\bibitem{tian2020fcos}
Z.~Tian, C.~Shen, H.~Chen, and T.~He, ``Fcos: A simple and strong anchor-free
  object detector,'' {\em IEEE Transactions on Pattern Analysis and Machine
  Intelligence}, 2020.

\bibitem{huang2019maskscore}
Z.~Huang, L.~Huang, Y.~Gong, C.~Huang, and X.~Wang, ``Mask scoring r-cnn,'' in
  {\em Proceedings of the IEEE/CVF Conference on Computer Vision and Pattern
  Recognition}, pp.~6409--6418, 2019.

\bibitem{wu2018group}
Y.~Wu and K.~He, ``Group normalization,'' in {\em Proceedings of the European
  conference on computer vision (ECCV)}, pp.~3--19, 2018.

\bibitem{kirillov2019panoptic}
A.~Kirillov, K.~He, R.~Girshick, C.~Rother, and P.~Doll{\'a}r, ``Panoptic
  segmentation,'' in {\em Proceedings of the IEEE/CVF Conference on Computer
  Vision and Pattern Recognition}, pp.~9404--9413, 2019.

\bibitem{wang2020axial}
H.~Wang, Y.~Zhu, B.~Green, H.~Adam, A.~Yuille, and L.-C. Chen, ``Axial-deeplab:
  Stand-alone axial-attention for panoptic segmentation,'' in {\em European
  Conference on Computer Vision}, pp.~108--126, Springer, 2020.

\bibitem{de2018jsis}
D.~de~Geus, P.~Meletis, and G.~Dubbelman, ``Panoptic segmentation with a joint
  semantic and instance segmentation network,'' {\em arXiv preprint
  arXiv:1809.02110}, 2018.

\bibitem{chen2019hybrid}
K.~Chen, J.~Pang, J.~Wang, Y.~Xiong, X.~Li, S.~Sun, W.~Feng, Z.~Liu, J.~Shi,
  W.~Ouyang, {\em et~al.}, ``Hybrid task cascade for instance segmentation,''
  in {\em Proceedings of the IEEE/CVF Conference on Computer Vision and Pattern
  Recognition}, pp.~4974--4983, 2019.

\bibitem{ke2021bcnet}
L.~Ke, Y.-W. Tai, and C.-K. Tang, ``Deep occlusion-aware instance segmentation
  with overlapping bilayers,'' in {\em Proceedings of the IEEE/CVF Conference
  on Computer Vision and Pattern Recognition}, pp.~4019--4028, 2021.

\bibitem{bolya2019yolact}
D.~Bolya, C.~Zhou, F.~Xiao, and Y.~J. Lee, ``Yolact: Real-time instance
  segmentation,'' in {\em Proceedings of the IEEE/CVF International Conference
  on Computer Vision}, pp.~9157--9166, 2019.

\bibitem{chen2019tensormask}
X.~Chen, R.~Girshick, K.~He, and P.~Doll{\'a}r, ``Tensormask: A foundation for
  dense object segmentation,'' in {\em Proceedings of the IEEE/CVF
  International Conference on Computer Vision}, pp.~2061--2069, 2019.

\bibitem{wang2020centermask}
Y.~Wang, Z.~Xu, H.~Shen, B.~Cheng, and L.~Yang, ``Centermask: single shot
  instance segmentation with point representation,'' in {\em Proceedings of the
  IEEE/CVF Conference on Computer Vision and Pattern Recognition},
  pp.~9313--9321, 2020.

\bibitem{lee2020centermask}
Y.~Lee and J.~Park, ``Centermask: Real-time anchor-free instance
  segmentation,'' in {\em Proceedings of the IEEE/CVF conference on computer
  vision and pattern recognition}, pp.~13906--13915, 2020.

\bibitem{wang2020solo}
X.~Wang, T.~Kong, C.~Shen, Y.~Jiang, and L.~Li, ``Solo: Segmenting objects by
  locations,'' in {\em European Conference on Computer Vision}, pp.~649--665,
  Springer, 2020.

\bibitem{wang2020solov2}
X.~Wang, R.~Zhang, T.~Kong, L.~Li, and C.~Shen, ``Solov2: Dynamic and fast
  instance segmentation,'' {\em Advances in Neural Information Processing
  Systems}, 2020.

\end{thebibliography}

\end{document}